\def\year{2022}\relax
\documentclass[letterpaper]{article} 
\usepackage{aaai22}  
\usepackage{times}  
\usepackage{helvet}  
\usepackage{courier}  
\usepackage[hyphens]{url}  
\usepackage{graphicx} 
\usepackage{natbib}  
\usepackage{caption} 
\DeclareCaptionStyle{ruled}{labelfont=normalfont,labelsep=colon,strut=off} 
\usepackage{amsfonts}

\usepackage[english]{babel}

\usepackage[title]{appendix}

\usepackage{comment}
\usepackage{color}
\usepackage{kotex}
\usepackage{adjustbox}

\usepackage{multicol}
\usepackage{multirow}
\usepackage{amsmath}
\usepackage{mathtools}
\usepackage{amssymb}
\usepackage{comment}
\usepackage[normalem]{ulem}

\usepackage{amsmath}
\usepackage{mathtools}
\usepackage{caption}
\usepackage{subcaption}
\usepackage{cancel}
\usepackage{caption}
\usepackage{subcaption}
\usepackage{url}

\usepackage{hyperref}       
\usepackage{url}            
\usepackage{booktabs}       
\usepackage{microtype}      %
\usepackage{xspace}

\usepackage{algorithm}
\usepackage{algpseudocode}

\usepackage{graphicx}
\graphicspath{{IMAGE/}}

\usepackage{newfloat}
\usepackage{listings}
\floatstyle{ruled}
\newfloat{listing}{tb}{lst}{}
\floatname{listing}{Listing}
\title{Dynamic Collective Intelligence Learning: Finding Efficient Sparse Model  \\ via Refined Gradients for Pruned Weights}
\author {
    Jangho Kim,
    Jayeon Yoo\equalcontrib, 
    Yeji Song\equalcontrib,
    KiYoon Yoo,
    Nojun Kwak
}
\affiliations {
    Seoul National University\\
    \{kjh91, jayeon.yoo, ldynx, 961230, nojunk\}@snu.ac.kr
}

\usepackage{bibentry}

\begin{document}

\maketitle

\begin{abstract}
With the growth of deep neural networks (DNN), the number of DNN parameters has drastically increased. This makes DNN models hard to be deployed on resource-limited embedded systems. To alleviate this problem, \textit{dynamic pruning} methods have emerged, which try to find diverse sparsity patterns during training by utilizing Straight-Through-Estimator (STE) to approximate gradients of pruned weights. STE can help the pruned weights revive in the process of finding dynamic sparsity patterns. However, using these coarse gradients causes training instability and performance degradation owing to the unreliable gradient signal of the STE approximation. In this work, to tackle this issue, we introduce refined gradients to update the pruned weights by forming dual forwarding paths from two sets (pruned and unpruned) of weights. We propose a novel Dynamic Collective Intelligence Learning (DCIL) which makes use of the learning synergy between the collective intelligence of both weight sets. We verify the usefulness of the refined gradients by showing enhancements in the training stability and the model performance on the CIFAR and ImageNet datasets. DCIL outperforms various previously proposed pruning schemes including other dynamic pruning methods with enhanced stability during training. 
\end{abstract}

\section{Introduction}
Massive improvements in various computer vision tasks using deep neural networks have been made at a cost of increase in millions of model parameters. While the task performances look promising, using such over-parameterized neural networks in low-end devices is infeasible due to demanding memory requirements and high latency. To resolve these issues, pruning unimportant units --  individual weights (unstructured pruning) and neurons (structured pruning) --  of neural networks has been well explored. 

Following its application to modern neural networks \citep{han2015learning}, many works have shown success through various pruning criteria in attaining a much sparse network that performs on par with or even better than the original unpruned network \cite{guo2016dynamic, xiao2019autoprune, renda2020comparing}. 
 
While effective, the aforementioned methods require a finetuning phase after the training phase and use a single fixed sparsity mask to obtain the pruned model. To tackle these problems, some works have focused on pruning methods \textit{during training} with dynamic sparsity patterns. Exploring diverse sparsity patterns is very important as it has a good chance of outperforming algorithms using a single sparsity pattern. These methods called \textit{dynamic pruning} discover sparse masks $(\mathbf{M})$ that are applied to the parameters of the neural network during training. Because fixing the masks discovered in the early training phase may have a detrimental effect on the training process, the masks and hence the sparsity pattern $(\mathbf{\overline{W}=M\odot W})$ are designed to dynamically change every few iterations ($\mathbf{F}$) by a certain criterion (e.g. L2 norm) on the real weights $(\mathbf{W})$ as the training progresses. 

\begin{figure}[t]
  \centering
   \includegraphics[width=1.0\linewidth]{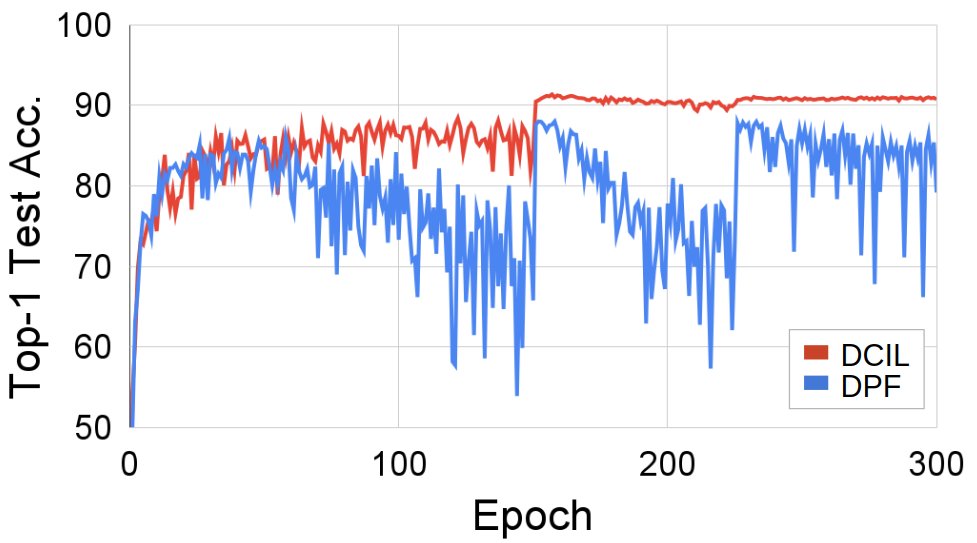}\\

  \caption{Test accuracy vs. epoch with ResNet-20 on CIFAR-10 by 95\% pruning. \textcolor{blue}{Blue}: Dynamic pruning with feedback (DPF) \cite{lin2020dynamic}, a state-of-the-art dynamic pruning method, \textcolor{red}{Red}: Dynamic collective intelligence learning (DCIL; Ours). DPF is unstable and shows a degradation in the performance because of coarse gradients. On the other hand, DCIL utilizes refined gradients to tackle the above issues. More details of DCIL and the stability analysis are stated in the \textit{Method} and \textit{Experiment} sections, respectively.}  \label{fig:stability}
\end{figure}

\begin{figure*}[t]
  \centering
   \includegraphics[width=1\linewidth]{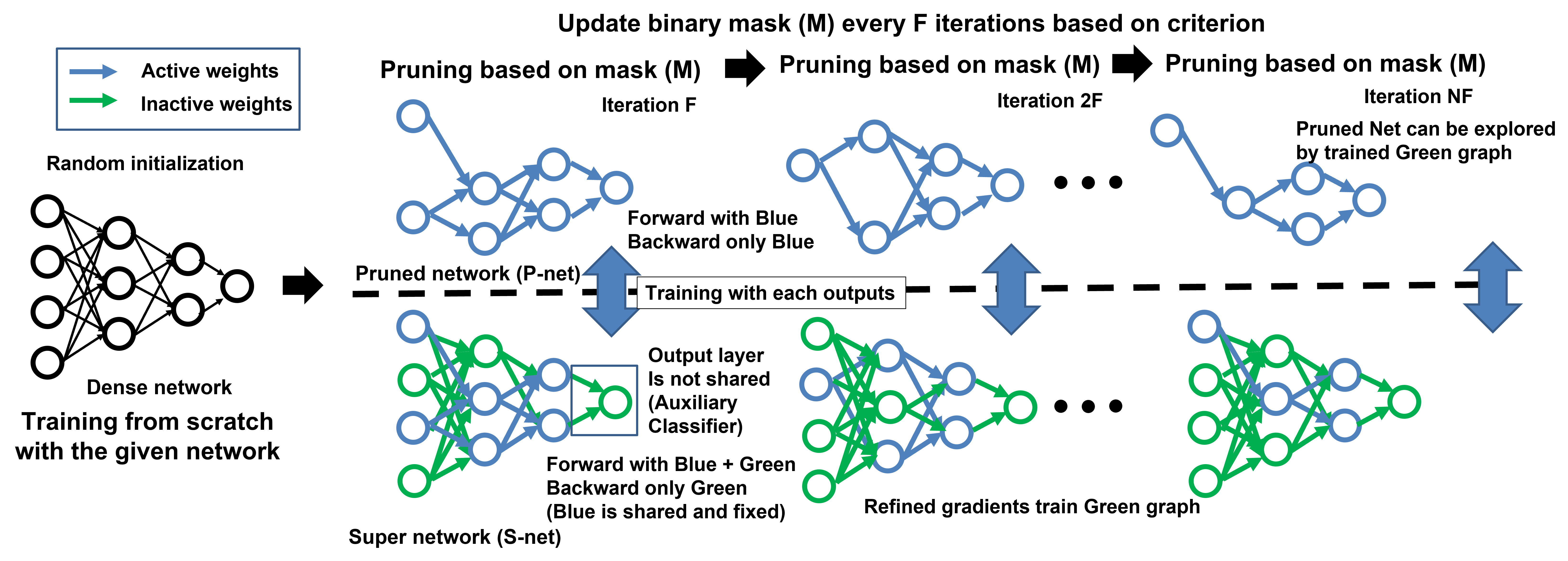}\\
  \caption{The overall process of Proposed DCIL. The blue and green graph refer to active and inactive weights, respectively. At all time, the S-net contains the P-net by sharing the weights of the P-net except a linear classifier and batch norm layers. Refined gradients make the inactive weights good candidates for future inclusion to the active weights.   }  \label{fig:overall}
\end{figure*}

Trivially, the gradient with respect to the weight of a pruned unit is zero as the pruned weight does not contribute anything to the output of the neural network nor the loss function ($\frac{\partial L}{\partial w}|_{m=0}=0$). However, to update the weights of the pruned unit, several works \citep{guo2016dynamic, lin2020dynamic} in \textit{dynamic pruning} have resorted to using a form of coarse gradient, in which the derivative of the masked weight with respect to the real weight is simply approximated to 1 (i.e $\frac{\partial \overline{w}}{\partial w}$=1). By doing so, the gradient $\frac{\partial L}{\partial {w}}$ is approximated by $\frac{\partial L}{\partial \overline{w}}$. This empirical practice has been applied in other literature such as quantization in the name of Straight-Through-Estimator (STE) \cite{bengio2013estimating}. The mask $\mathbf{M}$ is updated every $\mathbf{F}$ iterations and some of the inactive (pruned) weights  trained with coarse gradients may be revived to active (unpruned) weights. While using this form of coarse gradient is a convenient practice to revive the inactive weights, it may update the weights to an unintended direction harming the training stability and performance. This phenomenon is depicted in Fig. \ref{fig:stability} and  \ref{fig:stability2}.

 In this work, we show that using coarse gradient to update the inactive weights can be improved upon by forming a separate path for the inactive weights and updating them separately using an auxiliary loss function. Fig. \ref{fig:overall} shows the overall process of the proposed method. We form two different networks in size, called Super network (S-net) that has the entire set of real weights ($\mathbf{W}$) and Pruned network (P-net) that only utilizes the masked weights ($\mathbf{\overline{W}}$). 

We update the inactive weights (Green graph in Fig. \ref{fig:overall}) using the refined gradients of the S-net, whereas the same set of weights in the P-net is masked out to zero and not updated in back-propagation. Updating the inactive weights with refined gradients makes them qualified candidates of active weights for the ensuing training phase that can be readily used without destabilizing the task performance. 

A similar motivation, in which subsets of the neural network are stochastically selected at training for a better performance, is realized in Dropout \citep{JMLR:v15:srivastava14a} to incorporate the effect of model ensemble. In contrary, we update the inactive weights by back-propagation via the forward path of the S-Net for future inclusion in the set of active weights. 
Note that, the S-net is an unpruned model, whose weights are shared with the P-net, meaning negligible additional memory is needed for maintaining both the S-net and the P-net. In other words, the two networks are subnetworks of the original network whose structure is identical to the S-net. The S-net and P-net collaborate closely to find an efficient sparse network using refined gradients. Based on the learning synergy from the collective intelligence of both networks, we propose a novel sparse training framework named Dynamic Collective Intelligence Learning (DCIL). This allows even the inactive weights to obtain a much meaningful feedback than when using STE, which enhances training stability and achieves higher final performance in dynamic pruning as depicted in Fig. \ref{fig:stability}.

\section{Related work}

Earlier works in model pruning have focused on pruning unimportant parts of a fully trained model. \citet{lecun1990optimal}, and \citet{hassibi1993second} have used second derivative information to identify unimportant neurons. \citet{han2015learning} have established a general paradigm of iterative training-and-pruning to recover the degradation of accuracy due to pruning. Since then, many methods have been proposed in both unstructured pruning \citep{renda2020comparing, xiao2019autoprune}  and structured pruning \citep{he2017channel,li2016pruning}. \citet{guo2016dynamic} proposed a method to undo the pruning of important units by ``splicing" during the pruning process as an alternative to greedy pruning. This avoids permanent pruning of important units as units may revive in the splicing process.

Motivated by reducing the computational costs of iterative pruning methods, some recent methods have focused on finding a sparse network while training. Many methods dynamically prune and regrow pruned weights while training instead of using a fixed mask as done in earlier work of  \citet{guo2016dynamic}. Sparse Evolutionary Training \citep[SET]{mocanu2018scalable} uses simple heuristics to prune unimportant weights and subsequently regrow weights. Dynamic Sparse Reparameterization \citep[DSR]{mostafa2019parameter} adjusts sparsity patterns while training and automatically reallocates a sparsity level across layers. Sparse Momentum \citep[SM]{dettmers2019sparse} uses gradient momentum to determine unimportant weights and layers and redistributes sparsity across layers using the momentum. Dynamic Feedback with Pruning \citep[DPF]{lin2020dynamic} generalizes the simple scheme of \citet{guo2016dynamic} to prune while training and achieves state-of-the-art accuracy in unstructured pruning. While effective, the training curve is often unstable due to the coarse gradients, which makes model selection difficult for deployment. We propose a dynamic pruning method with superior final accuracy and is particularly stable when training by improving upon DPF via refining the gradients with two sub-networks.

\section{Proposed method}
In this section, we first describe incremental and dynamic pruning. Using incremental pruning, we gradually adjust the pruning sparsity ratio. We explain dynamic pruning method used in \citet{lin2020dynamic} to introduce our intuition of refined gradient. Then, we explain our proposed DCIL. 
\subsection{Backgrounds}
\noindent\textbf{Incremental pruning} One-shot pruning methods usually prune the network and finetune the pruned model for additional epochs to improve the performance. On the other hand, incremental pruning prunes the model each epoch by increasing sparsity ratio based on the current epoch. It is widely used in common pruning methods because it does not need an additional finetuning phase. In this work, we adopt the gradual pruning scheduling proposed in \citet{zhu2017prune} following the setting of \citet{lin2020dynamic}:

\begin{equation} 
S_c = S_t + (S_i-S_t)(1-\frac{c-c_0}{n})^3.
\label{eq:sparsity_ratio}
\end{equation} 
We gradually increase the sparsity ratio from an initial sparsity ratio ($S_i$ at $c_0=0$) based on the current epoch $c$ to the target sparsity ratio ($S_t$). $S_c$ means the current sparsity at $c$, where $c \in \{c_0, ..., c_0+n\}$ and $n$ is the number of epochs for the overall training.

\noindent\textbf{Dynamic pruning} We consider the binary mask $\mathbf{M}\in\{0, 1\}^P$ for obtaining the sparse weights by masking out the dense weights ($\mathbf{\overline{{W}}={M} \odot W}, \mathbf{W} \in \mathbb{R}^P$ where $P$ is the number of parameters and $\odot$ denotes the Hadamard product). A pruned network consists of these sparse weights ($\mathbf{\overline{{W}}}$). Dynamic pruning updates the binary mask $\mathbf{M}$ every $\mathbf{F}$ iterations based on a certain criterion such as the L2 norm and increases the sparsity with the incremental pruning scheme (\ref{eq:sparsity_ratio}).
The update rule of DPF \cite{lin2020dynamic} can be written by the following equation for a given loss function $\mathcal{L}$.  
\begin{equation}
    {w_i} \leftarrow {w_i} - \eta 
      \frac{\partial \mathcal{L}}{\partial \overline{w}_i}\frac{\partial \overline{w}_i}{\partial w_i},  \:\: \forall i \in \{1,\dots,P\} 
\label{Dynamic_STE}
\end{equation}
where $i$ is the index of weights and mask and $\overline{w}_i=m_i\times w_i.$ Sparse weights consist of two mutually exclusive sets which are active (unpruned) ($ \{w_i | w_i = \overline{w}_i, m_i=1\} $) and inactive (pruned) ($ \{w_i | w_i\neq \overline{w}_i, \ \overline{w}_i=m_i=0\} $).
Many dynamic pruning methods \cite{guo2016dynamic,lin2020dynamic} utilize the straight-through estimator \cite[STE]{bengio2013estimating} to flow the gradients to the inactive weights. Specifically, if the binary mask prunes the weight ($w_i$) as zero, the forward path is calculated by $\overline{w}_i=0$ and the backward path updates $w_i$. Thus, the gradient $\frac{\partial \overline{w}_i}{\partial w_i}$ of inactive weights is approximated as 1 using STE.

The instability in the test performance curve of DPF can be observed in Fig. \ref{fig:stability} and Fig. \ref{fig:stability2} as the masking pattern dynamically revives the ill-updated inactive weights.
In other words, the inactive weights updated with the STE-approximated gradient are candidates for being active weights based on a certain criterion. When ill-updated weights are selected, the pruned network is not prepared to perform well so it needs some iterations to recover its degraded accuracy. We tackle these issues by refining the gradients in the following section.

\begin{figure}[t]
  \centering
   \includegraphics[width=1\linewidth]{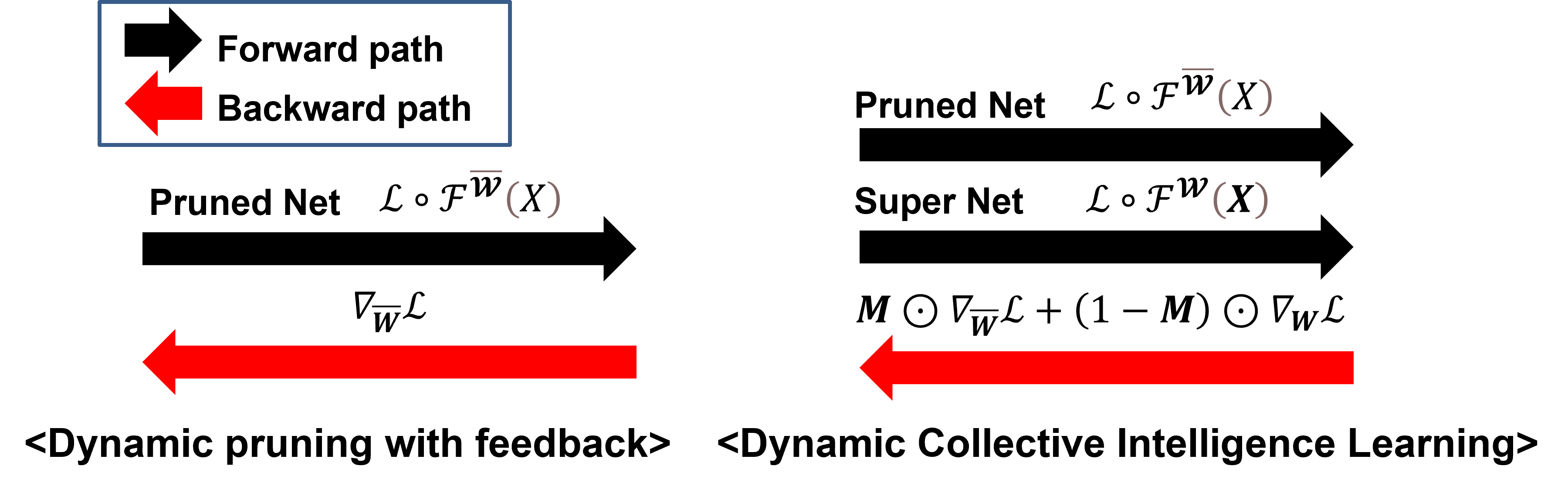}\\

  \caption{Comparison between DPF and DCIL of the forward and backward paths.}  \label{fig:forwarding_graph}
\end{figure}

\begin{table*}[t]
	\centering
	\caption{%
		Top-1 test accuracy of various SOTA pruning methods on \textbf{CIFAR-10} for unstructured weight pruning. The numbers of SM, DSR and DPF methods are from \citet{lin2020dynamic}. The $\star$ means that the model does not converge. $^\dagger$ indicates our numbers. DPF$^\dagger$ is from using the official code and we report the last and best accuracy of both DPF$^\dagger$ and our DCIL. We also report accuracy of the Dense$^\dagger$ models which are not pruned at all. All experiments were conducted three times.
	}
	\label{tab:sota_dnns_cifar10_unstructured_pruning_baseline_performance}
	\resizebox{0.95\textwidth}{!}{%
	\begin{tabular}{lccccccccc}
		\toprule
		                    						&                       				& \multicolumn{7}{c}{Methods}          																										& 		\\ \cmidrule{3-9}
	\multirow{2}{*}{\centering Model}   			& \multirow{2}{*}{\centering {Dense$^\dagger$}}		    & \multirow{2}{*}{\parbox{2cm}{\centering SM\\(DZ, \citeyear{dettmers2019sparse})}} 	& \multirow{2}{*}{\parbox{2cm}{\centering DSR\\(MW, \citeyear{mostafa2019parameter})}} & \multirow{2}{*}{\parbox{2cm}{\centering DPF\\(Lin, \citeyear{lin2020dynamic})}}
	& \multicolumn{2}{c}{\centering DPF$^\dagger$ (Lin, \citeyear{lin2020dynamic})} & \multicolumn{2}{c}{\centering DCIL~(Ours)} &
	\multirow{2}{*}{\centering {\shortstack[c]{Target \\ Pr. ratio}}}				    			
	\\ \cmidrule(lr){6-7} \cmidrule(lr){8-9} &&&&& {\centering Last} & {\centering Best} & {\centering Last} & {\centering Best} &
	\\ \midrule
		\multirow{2}{*}{ResNet-20}            		& \multirow{2}{*}{$92.36 \pm 0.10$} & $89.76 \pm 0.40$ & $87.88 \pm 0.04$ & $90.88 \pm 0.07$ & $88.02 \pm 0.12$ & $90.87 \pm 0.25$ & $\mathbf{91.61} \pm 0.20$ & ${91.93} \pm 0.11$ & 90\%      \\&& $83.03 \pm 0.74$ & $\star$ & $88.01 \pm 0.30$ & $81.46 \pm 2.26$	& $87.84 \pm 0.11$	& $\mathbf{90.54} \pm 0.19$	& ${90.80} \pm 0.23$ & 95\%                               	\\ \midrule
		\multirow{2}{*}{ResNet-32}            		& \multirow{2}{*}{$93.22 \pm 0.07$} & $91.54 \pm 0.18$ & $91.41 \pm 0.23$ &	$92.42 \pm 0.18$ & $91.14 \pm 0.12$	& $92.39 \pm 0.09$ & $\mathbf{93.05} \pm 0.15$ &  ${93.23} \pm 0.06$	& 90\% \\
        && $88.68 \pm 0.22$ & $84.12 \pm 0.32$ & $90.94 \pm 0.35$ & $86.52 \pm 1.38$ & $90.91 \pm 0.45$ & $\mathbf{92.04} \pm 0.24$ & ${92.27} \pm 0.27$ &   95\%             \\ \midrule
		\multirow{2}{*}{ResNet-56}            		& \multirow{2}{*}{$94.34 \pm 0.19$} & $92.73 \pm 0.21$ & $93.78 \pm 0.20$ & $93.95 \pm 0.11$ & $93.62 \pm 0.14$ & $93.97 \pm 0.14$ & $\mathbf{94.16} \pm 0.08$ & ${94.29} \pm 0.05$ &  90\% \\
        && $90.96 \pm 0.40$ & $92.57 \pm 0.09$ & $92.74 \pm 0.08$ & $90.59 \pm 0.38$  & $92.81 \pm 0.06$   & $\mathbf{93.75} \pm 0.14$	& ${93.98} \pm 0.12$	&  95\% \\ \midrule
		\multirow{3}{*}{WideResNet-28-2}            & \multirow{3}{*}{$94.73 \pm 0.03$} & $93.41 \pm 0.22$ & $93.88 \pm 0.08$ & $94.36 \pm 0.24$ & $94.08 \pm 0.34$  & $94.32 \pm 0.22$ & $\mathbf{94.69} \pm 0.13$	& ${94.84} \pm 0.28$ & 90\% \\
        && $92.24 \pm 0.14$ & $92.74 \pm 0.17$ & $93.62 \pm 0.05$ & $93.13 \pm 0.24$ & $93.61 \pm 0.08$	& $\mathbf{94.01} \pm 0.21$	& ${94.21} \pm 0.04$ & 95\% \\
        && $85.36 \pm 0.80$ & $\star$ & $88.92 \pm 0.29$ & $85.82 \pm 0.45$	& $88.77 \pm 0.23$	& $\mathbf{91.19} \pm 0.16$	& ${91.35} \pm 0.24$& 99\% \\ 
		\bottomrule
	\end{tabular}%
	}
    \vspace{-1em}
\end{table*}

\subsection{Dynamic Collective Intelligence Learning}
In this work, we propose Dynamic Collective Intelligence Learning (DCIL) which updates network by using the knowledge and refined gradients from two sub-networks. In the case of inactive weights, the gradient $\frac{\partial \overline{w}}{\partial w}$ should be approximated in order to update the inactive weights (pruned weights). To resolve this problem, DCIL calculates the refined gradient using an auxiliary classifier rather than relying on the approximated gradient ($\frac{\partial L}{\partial \overline{w}}$) by introducing dual forwarding paths corresponding to sub-networks. DCIL has two sub-networks, the Pruned network (P-net) and the Super network (S-net). 
The P-net (Blue graph in Fig. \ref{fig:overall}) consists of weights after pruning ($\mathbf{\overline{W}={M} \odot W}$) and the S-net (Blue + Green graph in Fig. \ref{fig:overall}) consists of the P-net and the complementary inactive weights ($\mathbf{W=(1-{M}) \odot W+M \odot W}$). Once pruning is done, the weight set of the S-net is a proper superset of that of the P-net (S-net $\supset$ P-net) because a set of inactive weights is a complementary set of active weights building the P-net ($\text{inactive weights} = \text{active weights}^C$). Note that the S-net contains the P-net and the size of the S-net is equal to that of the unpruned network. The forms of both the P-net and S-net vary every $\mathbf{F}$ iteration based on (\ref{eq:sparsity_ratio}) as depicted in Fig. \ref{fig:overall}.  

In contrary to the conventional dynamic pruning network, our DCIL does not use STE to update the inactive weights. We introduce a separate auxiliary classifier and batch norm layers for the S-net because the activation statistics are distinct throughout the forward paths of the S-net and P-net. The S-net computes the refined gradient of inactive weights with this auxiliary classifier. In this stage, the inactive weights collaborate with the P-net (active weights) to forward the output of the auxiliary classifier by sharing the weights of the P-net (See Fig. \ref{fig:overall}). The refined gradient helps the inactive weights to be prepared for the future inclusion to the P-net because the auxiliary classifier computes the refined gradient by forwarding with real weights ($\mathbf{W}$) rather than masked weights ($\mathbf{\overline{W}}$). In doing so, the inactive weights can be updated using the refined gradients with actual values of inactive weights. When the binary mask $\mathbf{M}$ changes every $\mathbf{F}$ iterations, inactive weights trained with refined gradients become better qualified candidates of the P-Net, which have better performance and stability than the inactive weights trained with approximated gradients with STE for active weights.

\begin{table*}[t]
	\centering
	\caption{%
		Top-1 test accuracy of our DCIL and DPF$^\dagger$ on \textbf{CIFAR-100} for unstructured weight pruning. $^\dagger$ indicates our numbers. DPF$^\dagger$ indicates our numbers using the official code and all reported numbers are averaged over three times. 
	}
	\label{tab:sota_dnns_cifar100_unstructured_pruning_baseline_performance}
	\vspace{-2mm}
	\resizebox{0.8\textwidth}{!}{%
	\begin{tabular}{lccccccc}
		\toprule
	\multirow{2}{*}{\centering Model}   			&
	\multirow{2}{*}{\parbox{2cm}{\centering Dense$^\dagger$}}
		\!\!\!& \multicolumn{2}{c}{\centering DPF$^\dagger$} & \multicolumn{2}{c}{\centering DCIL (Ours)}		& \multirow{2}{*}{\parbox{2cm}{\centering {Target \\ Pr. ratio}}}				    			
	\\ \cmidrule(lr){3-4} \cmidrule(lr){5-6} && {\centering Last} & {\centering Best} & {\centering Last} & {\centering Best} &
	\\ \midrule
		\multirow{2}{*}{ResNet-20} & \multirow{2}{*}{$67.81 \pm 0.54$}				& $53.7 \pm 1.95$ & $63.94 \pm 0.02$ & $\mathbf{67.12} \pm 0.06$ & ${67.51} \pm 0.10$ & 90\%                               	\\ 
				            						& & $46.15 \pm 0.16$	& $58.17 \pm 0.46$	& $\mathbf{64.10} \pm 0.53$	& ${64.68} \pm 0.37$	& 95\%                               	\\ \midrule
		\multirow{2}{*}{ResNet-32}            		& \multirow{2}{*}{$69.50 \pm 0.20$} 	& $61.12 \pm 1.73$	& $67.82 \pm 0.26$ & $\mathbf{69.96} \pm 0.28$ &  ${70.40} \pm 0.36$ & 90\% \\ & & $50.08 \pm 6.10$ & $63.22 \pm 0.39$ & $\mathbf{67.90} \pm 0.14$ & ${68.53} \pm 0.16$ &  95\%                          		\\ \midrule
		\multirow{2}{*}{ResNet-56}            		& \multirow{2}{*}{$74.23 \pm 0.13$} 	& $67.99 \pm 1.15$ & $72.99 \pm 0.92$ & $\mathbf{74.59} \pm 0.36$	& ${74.95} \pm 0.24$ & 90\% \\ && $56.88 \pm 0.85$  & $69.83 \pm 0.42$   & $\mathbf{73.30} \pm 0.34$	& ${73.63} \pm 0.39$	& 95\%                      			\\ \midrule
		\multirow{3}{*}{WideResNet-28-2}            & \multirow{3}{*}{$74.81 \pm 0.15$} 		& $71.32 \pm 0.59$  & $73.45 \pm 0.25$    & $\mathbf{74.61} \pm 0.02$	& ${74.97} \pm 0.13$ & 90\% 	\\ && $68.56 \pm 0.50$	& $71.69 \pm 0.36$	& $\mathbf{73.70} \pm 0.23$	& ${74.05} \pm 0.13$	& 95\%                          		\\ && $51.27 \pm 0.33$	& $60.02 \pm 0.04$	& $\mathbf{65.55} \pm 0.18$	& ${66.00} \pm 0.10$	&99\%  \\ 
		\bottomrule
	\end{tabular}%
	}
    \vspace{-1em}
\end{table*}

\begin{figure}[t]
  \centering
   \includegraphics[width=1\linewidth]{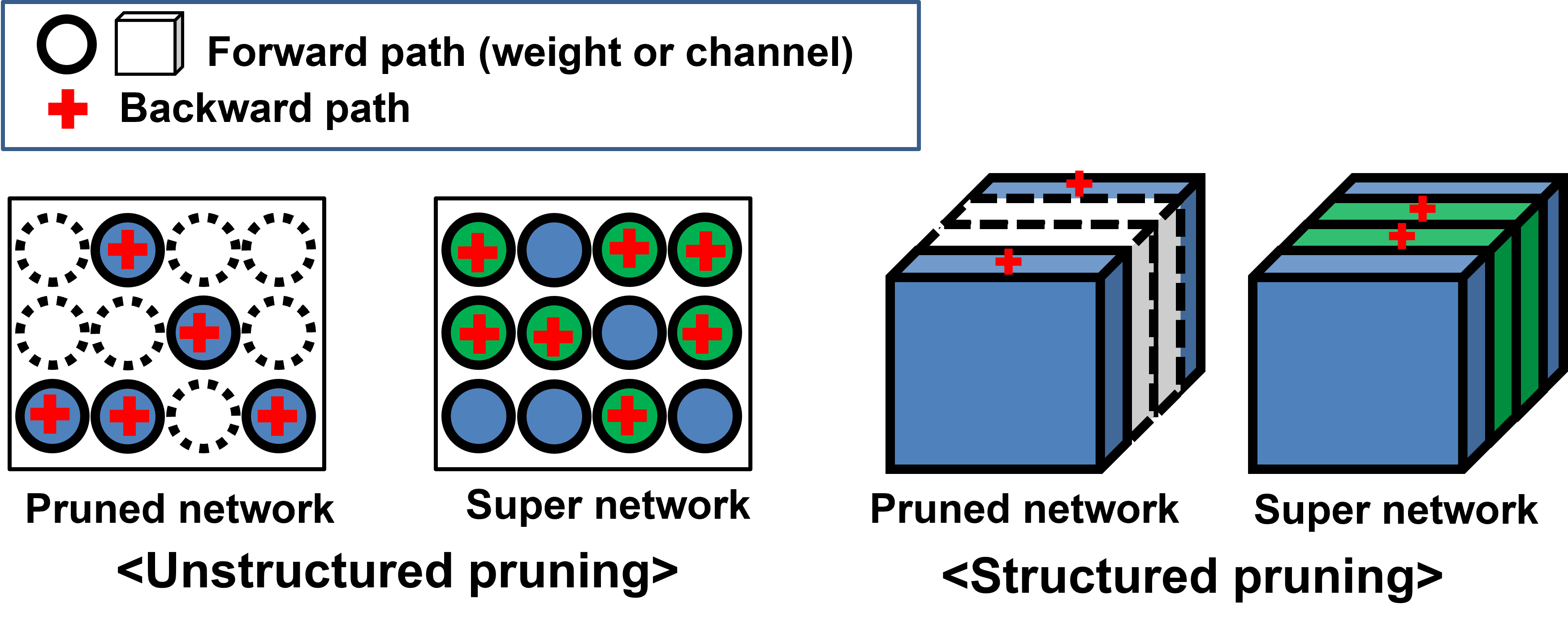}\\

  \caption{Diagram of applying DCIL to two pruning types, unstructured pruning and structured pruning.}  \label{fig:pruning_type}
\end{figure}

Let $\mathcal{F}^\mathbf{W}: \mathcal{X} \rightarrow \mathcal{Y}$ and $\mathcal{L}: \mathcal{Y} \rightarrow \mathbb{R}$ be the forwarding path with given weights ($\mathbf{W}$) and the loss function, respectively. Then, we can define the update rule of DCIL as follows:

\begin{equation}
\begin{split}
      &\mathbf{W}  \leftarrow \mathbf{W} - \eta\{\mathbf{M} \odot \nabla_{\overline{\mathbf{W}}}\mathcal{L}
            +(1-\mathbf{M}) \odot \nabla_{{\mathbf{W}}}\mathcal{L}\} \\
    & \nabla_{\overline{\mathbf{W}}}\mathcal{L} \triangleq \frac{\partial \mathcal{L} \circ \mathcal{F}^\mathbf{\overline{W}}( \mathbf{X})}{\partial \overline{\mathbf{w}}}, \quad 
    \nabla_{{\mathbf{W}}}\mathcal{L} \triangleq
    \frac{\partial \mathcal{L} \circ \mathcal{F}^\mathbf{{W}}( \mathbf{X})}{\partial {\mathbf{w}}}.
\end{split}
\label{Dynamic_STE}
\end{equation}
In this setting, note that the composite functions $\mathcal{L} \circ \mathcal{F}^\mathbf{\overline{W}}( \mathbf{X})$ and $\mathcal{L} \circ \mathcal{F}^\mathbf{W}( \mathbf{X})$ correspond to the model with different classifiers of which the latter uses the auxiliary classifier. Fig. \ref{fig:forwarding_graph} shows the forward and  backward paths according to each algorithm. DPF \cite{lin2020dynamic} updates $\mathbf{W}$ with the gradient of $\mathbf{\overline{W}}$ by the approximation using STE ($\nabla_{\overline{\mathbf{W}}}\mathcal{L}$). On the other hand, DCIL updates $\mathbf{W}$ with gradients of $\mathbf{\overline{W}}$ and the refined gradients of inactive weights without an approximation ($\mathbf{M} \odot \nabla_{\overline{\mathbf{W}}}\mathcal{L}
            +(1-\mathbf{M}) \odot \nabla_{{\mathbf{W}}}\mathcal{L}$).
Generally, the cross entropy term $\mathcal{L}_{ce}$ is used for the loss function. However, we want separate paths to share more information then collaborate better with each other. Since DCIL has two forward paths with two classifiers, DCIL can expand the cross entropy loss to Knowledge distillation (KD) loss \cite{hinton2015distilling,NEURIPS2018_6d9cb7de} by using each output as a soft target for each Kullback–Leibler (KL) divergence term ($\textit{KL}(p||q; \mathcal{T})$) with a temperature value $\mathcal{T}$:   
\begin{equation} 
\resizebox{.9\hsize}{!}{$
\begin{split}
&\mathcal{L} \circ \mathcal{F}^\mathbf{\overline{W}}( \mathbf{X}) = \mathcal{L}_{ce} \circ \mathcal{F}^\mathbf{\overline{W}}( \mathbf{X}) + \lambda \mathcal{T}^2\cdot{\textit{KL}}( \mathcal{F}^\mathbf{{W}}( \mathbf{X})|| \mathcal{F}^\mathbf{\overline{W}}( \mathbf{X}); \mathcal{T}) \\
&\mathcal{L} \circ \mathcal{F}^\mathbf{{W}}( \mathbf{X}) = \mathcal{L}_{ce} \circ \mathcal{F}^\mathbf{{W}}( \mathbf{X}) + \lambda \mathcal{T}^2\cdot{\textit{KL}}( \mathcal{F}^\mathbf{\overline{W}}( \mathbf{X})|| \mathcal{F}^\mathbf{{W}}( \mathbf{X}); \mathcal{T})
\end{split}
$}
\label{eq:Supernet_loss}
\end{equation}

where $\lambda$ is a hyper-parameter for balancing loss functions. We compare the effect of the KL loss and the cross entropy loss in our ablation study. 

DCIL can be applied regardless of the pruning type, i.e, it can be applied to both unstructured or structured pruning. In the unstructured pruning case, inactive and active weights coexist in a filter so the pruning is done in the weight-level whereas DCIL of structured pruning operates in the filter-level (neuron) as shown in Fig. \ref{fig:pruning_type}.

\begin{table*}[t]
	\centering
	\caption{Top-1 and Top-5 test accuracy of ResNet-18 and ResNet-50 on \textbf{ImageNet}. DPF$^\dagger$ indicates our numbers using the official code. We report the best accuracy and the accuracy from the last epoch. Difference refers to difference between pruned and dense accuracy (`Last - Dense'). Best accuracy of Top-5 is selected based on Top-1 best epoch.}
	\label{tab:resnet50_imagenet_unstructured_pruning_baseline_performance}
	\vspace{-2mm}
	\resizebox{0.9\textwidth}{!}{%
	\begin{tabular}{lcccccccccc}
		\toprule
								\multirow{2}{*}{Model}				&		\multirow{2}{*}{Method}		 &\multicolumn{4}{c}{Top-1 accuracy} 														& \multicolumn{4}{c}{Top-5 accuracy}          												& \multirow{2}{*}{\parbox{2cm}{\centering {Target \\ Pr. ratio}}}										\\ \cmidrule(lr){3-6} \cmidrule(lr){7-10} 
		 & 												& Dense						& Last						& Best					& Difference &Dense						& Last 						& Best & Difference 					& 																	 \\ \midrule

	\multirow{5}{*}{ResNet-18}	&DPF$^\dagger$~(Lin, \citeyear{lin2020dynamic})       	& 70.50					& 69.08							& 69.19							& -1.31						& 89.54					& 88.93							& 88.88										& -0.66								& 80\%									\\
		
			&DCIL (Ours)       	& 	70.50					& \textbf{69.57}							& 69.62							& -0.88						& 89.54 						& \textbf{89.26}							& 89.29										& -0.25								& 80\%									\\
			\cmidrule{2-11}
		
			&DPF$^\dagger$~(Lin, \citeyear{lin2020dynamic})       	& 70.50						& {67.37}							& 67.66							& -2.83						& 89.54 						& {87.83}							& 87.88										& -1.66								& 90\%									\\
		
			&DCIL~(Ours)        	& 	70.50					& \textbf{68.66}							& 68.66							& -1.84						& 89.54 						& \textbf{88.57}							& 88.57										& -0.97								& 90\%	\\

			&DCIL w/o KL (Ours; $\lambda=0$)       	& 70.50						& 68.37							&68.41							& -2.09						& 89.54 						& 88.32							& 88.26										& -1.28								& 90\%	
			\\ \midrule

	\multirow{13}{*}{ResNet-50}	&Incremental~(ZG, \citeyear{zhu2017prune})       	& 75.95	&	--				& 74.25							& -1.70							& 92.91 & --						& 91.84 						& -1.07							& 80\%																										\\

		&DSR~(MW, \citeyear{mostafa2019parameter})	    	& 74.90				& --		& 73.30 						& -1.60							& 92.40	 & --					& 92.40 						& {0}					& 80\%																		 			\\
		&SM (DZ, \citeyear{dettmers2019sparse})              & 75.95			&	--		& 74.59							& -1.36							& 92.91		& --				& 92.37 						& -0.54							& 80\%											\\
		&DPF~(Lin, \citeyear{lin2020dynamic})                								& 75.95			&	--		& 75.48							& {-0.47}				& 92.91 	& --				& 92.59 						& -0.32							& 80\%									\\
		&DPF$^\dagger$~(Lin, \citeyear{lin2020dynamic})            								& 76.06					& 75.59							& 75.61				& -0.47						& 92.85							& 92.66				& 92.71										& -0.19								& {80\%}						 			\\
		
		&DCIL~(Ours)                								& 76.06							& \textbf{76.16}							& 76.17				& 0.10 					& 92.86 						& \textbf{92.94} & 93.00 & 0.09							& 80\%									\\

\cmidrule{2-11}

		&Incremental (ZG, \citeyear{zhu2017prune})       	& 75.95		&		--		& 73.36							& -2.59							& 92.91	 & --					& 91.27 						& -1.64							& 90\%																						 			\\
		&DSR (MW, \citeyear{mostafa2019parameter})     		& 74.90			 & --			& 71.60							& -3.30							& 92.40				 & --		& 90.50 						& -1.90							& 90\%					\\
		&SM (DZ, \citeyear{dettmers2019sparse})              & 75.95			 & --			& 72.65							& -3.30							& 92.91			& --			& 91.26 						& -1.65	& 90\%								 			\\
		&DPF~(Lin, \citeyear{lin2020dynamic})                								& 75.95					& --	& 74.55							& {-1.44}				& 92.91		& --				& 92.13							& {-0.78}& {90\%}						 			\\
		&DPF$^\dagger$~(Lin, \citeyear{lin2020dynamic})            								& 76.06					& 74.39							& 74.48				& -1.67						& 92.85							& 92.08				& 92.05										& -0.77								& {90\%}						 			\\
		&		DCIL~(Ours)                								& 76.06						& \textbf{75.34}							& 75.40				& -0.72						& 92.85							& \textbf{92.58}				& 92.63										& -0.27								& {90\%}						 			\\
		
		&		DCIL w/o KL (Ours; $\lambda=0$)               								& 76.06						& 74.75							& 74.87				& -1.31						& 92.85							& 92.31				& 92.24										& -0.54								& {90\%}						 			\\

		\bottomrule
	\end{tabular}%
	}
\end{table*}

\begin{table*}[t]
	\centering
	\caption{Top-1 test accuracy of our DCIL and other baseline methods on \textbf{CIFAR-10} for structured weight pruning. $^\dagger$ indicates our numbers. We ran DPF$^\dagger$ using the official code and official training schedule. Note that for a given pruning ratio DCIL and DPF prune filters across layers, meanwhile SFP prunes filters within the layer. Furthermore, we report the differences (`Pruned - Dense') in the Appendix. When comparing the differences, DCIL outperform with the higher gap in most cases.  All reported numbers are averaged over three times.}
	\label{tab:sota_dnns_cifar10_structured_pruning_performance}
	\resizebox{0.9\textwidth}{!}{%
	{
	\begin{tabular}{lcccccccc}
		\toprule
		\multirow{2}{*}{\parbox{2cm}{\centering Model}} & \multirow{2}{*}{\parbox{2cm}{\centering Dense$^\dagger$}} & \multirow{2}{*}{\parbox{1.5cm}{\centering SFP\\(H$^+$,\citeyear{he2018soft})}} & \multirow{2}{*}{\parbox{1.5cm}{\centering DPF\\(Lin,\citeyear{lin2020dynamic})}} & \multicolumn{2}{c}{\centering DPF$^\dagger$} & \multicolumn{2}{c}{\centering DCIL (Ours)} & \multirow{2}{*}{\parbox{2cm}{\centering Target Pr.\\ ratio}} \\
		\cmidrule(lr){5-6} \cmidrule(lr){7-8}
		& & & & \centering Last & \centering Best & \centering Last & \centering Best & \\
		\midrule
		\multirow{2}{*}{ResNet-32} & \multirow{2}{*}{$93.22 \pm 0.07$} & $92.07 \pm 0.22$ & $ 92.18 \pm 0.16 $ & $91.61 \pm 0.15$ & $91.81 \pm 0.19$ & $\mathbf{93.02} \pm 0.21$ & ${93.10} \pm 0.21$ & 30\% \\
		& & $ 91.14 \pm 0.45$ & $ 91.50 \pm 0.21 $ & $90.01 \pm 0.40$ & $90.28 \pm 0.40$ & $\mathbf{92.61} \pm 0.07$ & ${92.76} \pm 0.16$ & 40\% \\
		\midrule
		\multirow{2}{*}{WideResNet-28-2} & \multirow{2}{*}{$94.73 \pm 0.03$} & $ 94.02 \pm 0.24 $ & $ 94.52 \pm 0.08 $ & $94.11 \pm 0.17$ & $94.30 \pm 0.09$ & $\mathbf{94.55} \pm 0.04$ & ${94.61} \pm 0.08$ & 40\% \\
		& & $ 86.00 \pm 1.09 $ & $ 90.53 \pm 0.17 $ & $88.99 \pm 0.45$ & $89.76 \pm 0.64$ & $\mathbf{92.10} \pm 0.11$ & ${92.32} \pm 0.15$ & 80\% \\
		\midrule
		\multirow{2}{*}{WideResNet-28-4} & \multirow{2}{*}{$95.50 \pm 0.07$} & $ 95.15 \pm 0.11 $ & $ 95.50 \pm 0.05 $ & $95.36 \pm 0.16$ & $95.52 \pm 0.14$ & $95.46 \pm 0.06$ & ${95.59} \pm 0.06$ & 40\% \\
		& & $ 91.88 \pm 0.59 $ & $ 93.79 \pm 0.09 $ & $93.41 \pm 0.07$ & $93.51 \pm 0.11$ & $\mathbf{94.18} \pm 0.04$ & ${94.38} \pm 0.08$ & 80\% \\
		\midrule
		\multirow{2}{*}{WideResNet-28-8} & \multirow{2}{*}{$95.93 \pm 0.07$} & $ 95.62 \pm 0.04 $ & $ {96.06} \pm 0.12 $ & $95.86 \pm 0.17$ & $96.03 \pm 0.18$ & $95.73 \pm 0.07$ & $95.91 \pm 0.10$ & 40\% \\
		& & $ 94.22 \pm 0.21 $ & $ 95.15 \pm 0.03 $ & $94.97 \pm 0.20$ & $95.16 \pm 0.12$ & $\mathbf{95.43} \pm 0.09$ & ${95.49} \pm 0.05$ & 80\% \\
		\bottomrule
	\end{tabular}%
	}}
\end{table*}

\section{Experiment}
We compare our DCIL with the state-of-the-art methods on commonly used datasets. Our model significantly improves the performance and the stability in both unstructured and structured pruning regardless of the architecture.

We also show the stability analysis and the effect of the KL divergence through an ablation study. As a result of the ablation study, we demonstrate that updating the inactive weights with refined gradients 
plays a key role in improving performance and training stability than using KL divergence.

\subsection{Experiment Setting}

\subsubsection{Datasets}
We conduct experiments on three image classification datasets CIFAR-10, CIFAR100 \cite{krizhevsky2009learning}, and ImageNet \cite{ILSVRC15}.
\subsubsection{Baselines} We compare our method against DPF \citep{lin2020dynamic}, SM \citep{dettmers2019sparse},  DSR \citep{mostafa2019parameter} and SFP \citep{he2018soft}. Please refer to the section on \textit{Related Work} for the details.
\subsubsection{Implementation Details}
For a fair comparison, we followed the same experimental settings with DPF \cite{lin2020dynamic} for all experiments. We set the mask update frequency ($\mathbf{F}$) to 16 iterations like DPF.
We applied our pruning method and other pruning methods to ResNet \cite{he2016deep} and WideResNet \cite{zagoruyko2016wide}. Following DPF, we performed pruning across all convolutional layers except for the batch normalization layers and the last fully connected layer. We used SGD with the Nesterov momentum of 0.9. For CIFAR, the ResNet models were trained for 300 epochs with the initial learning rate of 0.2 and the L2 weight decay parameter of 1e-4. We decayed the learning rate by 10 at 150 and 225 epochs. WideResNet models were trained for 200 epochs with the initial learning rate of 0.1 which is decayed by 5 at 60, 120 and 160 epochs. The weight decay parameter was set to 5e-4. We used 128 mini-batch size for all experiments of CIFAR. In the ImageNet training, we trained models with 90 epochs and decayed the learning rate by 10 at 30, 60 and 80 epochs with the initial learning rate of 0.1. We used 128 mini-batch size for ResNet-18 and 1,024 mini-batch size for ResNet-50 following \cite{goyal2017accurate}. We introduce the warm-up epoch, an early period of learning without KL loss but only using cross-entropy ($\lambda=0$), because regularizing the model with KL loss using an untrained teacher model may hinder convergence. We set the warm-up epoch to 70 and 10 for CIFAR and ImageNet datasets, respectively. However, we confirmed that the scale of warm-up epoch does not significantly affect the test accuracy as shown in the Appendix. We set the $\lambda=1$ and $\mathcal{T}=2$ for the KD loss. We used the Pytorch framework for all experiments. All experiments on CIFAR and the experiments of ResNet-18 on ImageNet were conducted with a GeForce GTX 1080 Ti GPU. The experiments of ResNet-50 on ImageNet were conducted with four NVIDIA RTX A6000 GPUs.

\subsection{Experiment Results}

In all experiments, we report the numbers with same target pruning ratio from the original paper of DPF \cite{lin2020dynamic} and used the same L2 pruning criterion as DPF. We also re-implemented DPF referred as DPF$^\dagger$ to compare training stability. We report the test accuracy of the last and the best epoch of DPF and DCIL. Our last epoch accuracy is made bold if our number is higher than those of all the methods.

\subsubsection{Unstructured}
Table \ref{tab:sota_dnns_cifar10_unstructured_pruning_baseline_performance} shows the top-1 test accuracy of various SOTA pruning methods on the CIFAR-10 dataset. Along with the accuracy reported in the original papers of each method, we report the performance of DPF we re-implemented. 
DCIL significantly outperforms all other methods for all models, datasets, and a variety of pruning sparsities. In particular, the test accuracy in the last epoch of DCIL is 1-3\% higher than the best accuracy of re-implemented DPF which is nearly the same as the performance of the original paper. 
DCIL has a far superior performance on the CIFAR-100 dataset as shown Table \ref{tab:sota_dnns_cifar100_unstructured_pruning_baseline_performance}. The last accuracy of DCIL outperforms the last accuracy of DPF by 3-18\%, the best accuracy of DPF by 1-6\%, and sometimes the performance of the dense model due to the regularization effect.
In addition, the training instability of DPF is severe in the CIFAR datasets on unstructured pruning, resulting in a very large difference between the last and the best accuracies as shown in Table \ref{tab:sota_dnns_cifar10_unstructured_pruning_baseline_performance} and Table \ref{tab:sota_dnns_cifar100_unstructured_pruning_baseline_performance}. On the other hand, our DCIL has a stable training curve, demonstrated by the small margin between the last accuracy and the best accuracy. And as a result, 
in the absence of an additional fine-tuning phase or validation sets, DCIL has a good chance to work as good as the optimal pruned model even on datasets where training can be relatively unstable.

Also, we conducted ImageNet experiments to verify the effectiveness of DCIL in large-scale datasets as shown in Table \ref{tab:resnet50_imagenet_unstructured_pruning_baseline_performance}. DCIL outperforms other pruning methods with a large margin in test accuracy. In ResNet-18 with 90\% pruning, the performance gap between DPF and DCIL is around 1.3\%. The tendency of stability during training is similar to that of CIFAR, which is depicted in Appendix. We also report the results of ablation studies with respect to the KL divergence term. `DCIL w/o KL' in Table \ref{tab:resnet50_imagenet_unstructured_pruning_baseline_performance} means a model trained with DCIL by only cross-entropy. KL helps improve the performance by a range of 0.3-0.6\%. However, DCIL without KL also surpasses DPF by about 0.4-1\%, 
which shows the effectiveness of the refined gradients. A similar tendency is observed in the following stability analysis section.

\subsubsection{Structured} 
To show a broad applicability of our algorithm, we applied DCIL at the filter level. Unlike unstructured pruning, filter pruning or \emph{structured pruning} could take full advantage of the BLAS library and accelerate model inference. 
Table \ref{tab:sota_dnns_cifar10_structured_pruning_performance} shows a comparison of different methods on CIFAR-10 for ResNet and WideResNet variants. Both DPF and SFP dynamically allocate sparsity pattern, updating the filter with a coarse gradient ($\frac{\partial L}{\partial \overline{w}}$). 
DCIL outperforms all the baselines in most settings except for WideResNet-28-8 (40\% sparsity), where DCIL is comparable to the official results of DPF. DCIL's superior performance for structured pruning can be seen in terms of architecture search as structured pruning is considered as performing implicit architecture search \citep{liu2018rethinking}. DCIL searches for better subnetwork candidates obtained through the refined gradients, whereas existing methods using coarse gradients could be updated in a direction that is harmful to the performance and reach a suboptimal solution.

We also report the full results on CIFAR-100 with other structured pruning methods \citep{he2018soft, dong2017less, he2019filter} in the Appendix. Briefly, when compared with the existing methods for ResNet-56 (40\% and 50\% sparsity), DCIL shows higher accuracies in all settings. This result validates the effectiveness of our DCIL's refined gradient for structured pruning.

\begin{table}[t]
	\centering
	\caption{%
		\small
		Standard Deviation of the top-1 test accuracy over the last 10\% epochs. To compare the training stability of our DCIL and DPF on \textbf{CIFAR-10}, we report the standard deviation of the top-1 test accuracy over the last 10\% epochs (i.e. 30 epochs for ResNet and 20 epochs for WideResNet), which is expected to be stable with training for several epochs after the last learning rate decay. The results are averaged over three runs.
	}
	\label{tab:training_stability_cifar10}
	\resizebox{0.45\textwidth}{!}{%
	\begin{tabular}{lcccc}
		\toprule
		\multirow{2}{*}{\parbox{2cm}{\centering Model}} &
		\multicolumn{3}{c}{CIFAR-10} &
		\multirow{2}{*}{\parbox{2cm}{\centering Target \\Pr.ratio}} \\ \cmidrule(lr){2-4} & \parbox{2cm}{\centering Dense} & \centering DPF & \centering DCIL (Ours)		&
\\ \midrule
		\multirow{2}{*}{ResNet-20} & \multirow{2}{*}{$0.070 \pm 0.012$}				& $1.179 \pm 0.310$ & $0.072 \pm 0.013$ & 90\%                               	\\ && $3.839 \pm 1.455$ & $0.114 \pm 0.007$ & 95\%                               	\\ \midrule
		\multirow{2}{*}{ResNet-32} & \multirow{2}{*}{$0.061 \pm 0.013$}				& $0.344 \pm 0.036$ & $0.080 \pm 0.005$ & 90\%                               	\\ && $1.947 \pm 0.362$ & $0.097 \pm 0.011$ & 95\% \\ 
		\midrule
		\multirow{2}{*}{ResNet-56} & \multirow{2}{*}{$0.060 \pm 0.009$}				& $0.179 \pm 0.055$ & $0.063 \pm 0.007$ & 90\%                               	\\ && $0.990 \pm 0.078$ & $0.082 \pm 0.009$ & 95\% \\
		\midrule
		\multirow{3}{*}{WideResNet-28-2} & \multirow{3}{*}{$0.061 \pm 0.004$}	& $0.120 \pm 0.033$ & $0.093 \pm 0.011$ & 90\%  
		\\ && $0.286 \pm 0.067$ & $0.086 \pm 0.015$ & 95\% 
		\\ && $2.790 \pm 0.723$ & $0.096 \pm 0.007$ & 99\% \\
				\bottomrule
	\end{tabular}%
	}
    \vspace{-1em}
\end{table}

\begin{figure*}[t]
\centering
  \begin{subfigure}[b]{0.4\textwidth}
    \includegraphics[width=1\textwidth]{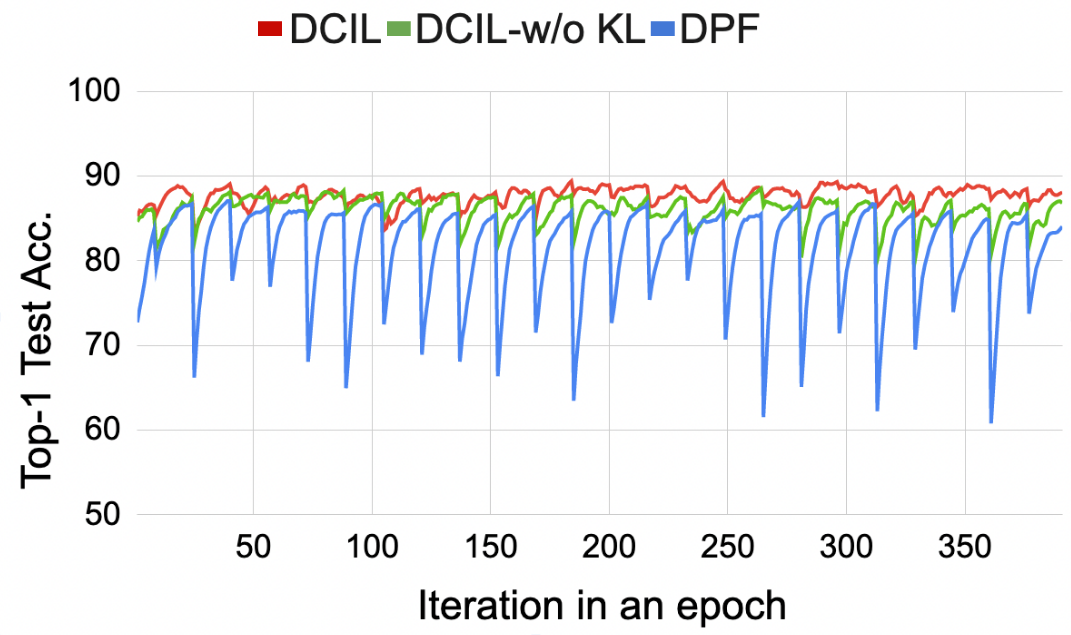}
  \caption{120th epoch}
  \end{subfigure}
  \centering
  \begin{subfigure}[b]{0.4\textwidth}
    \includegraphics[width=1\textwidth]{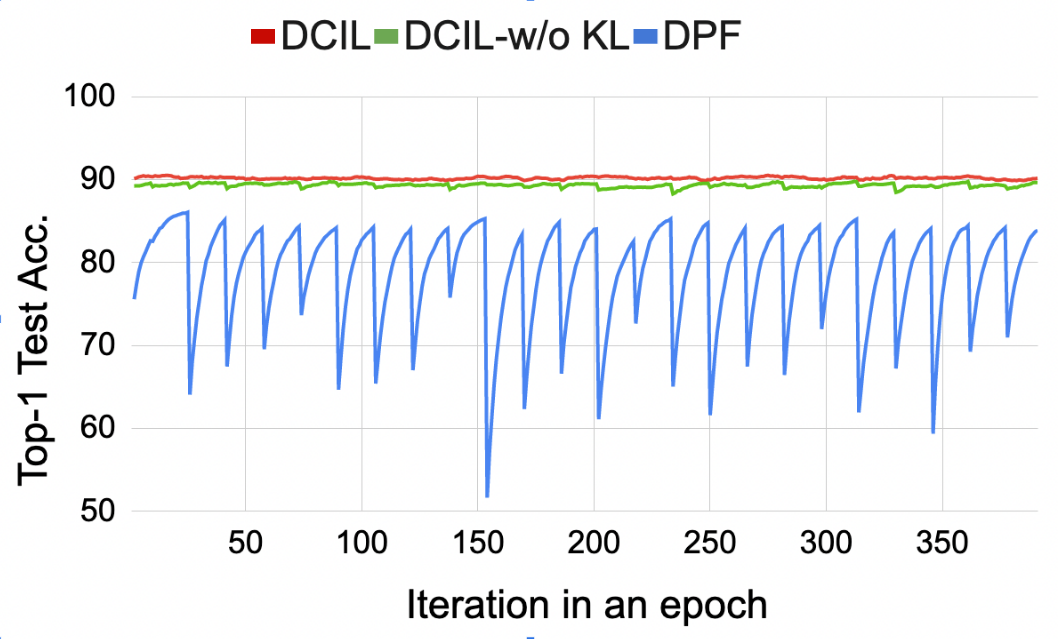}
    \caption{200th epoch}
  \end{subfigure}
  \caption{Top-1 test accuracy of every iteration in an epoch. To show the effectiveness of changes in active weights caused by changes in pruning masks, we evaluated ResNet-20 pruned with 95\% sparisty using \textcolor{red}{Red}: DCIL, \textcolor{green}{Green}: DCIL without KL loss, and \textcolor{blue}{Blue}: DPF on CIFAR10 every iteration at (a) 120th epoch and (b) 200th epoch. The accuracy of all methods shows a repetitive pattern with a pruning frequency ($\mathbf{F}$) of 16.}
  \label{fig:stability2}
\end{figure*}

\subsection{Analysis of the stability}\label{ablation}
Table \ref{tab:training_stability_cifar10} summarizes the standard deviation of top-1 test accuracy of ours and DPF over the last 30 and 20 epochs, which is the last 10\% epochs for ResNet and WideResNet, respectively. Despite the low learning rate since it is after the last decaying epoch, DPF has a large variance in top-1 test accuracy, while our method has a much smaller variance. We conjecture that this is due to the differences in the gradients used to update the inactive weights, i.e, our method updates inactive weights using refined gradients calculated on S-net, whereas DPF approximately estimate gradients that can lead to a degradation of the pruned model.

In order to demonstrate the effect of using refined gradient to dynamically allocate sparsity masks on performance directly, we evaluated the pruned model using DCIL, DCIL without KL, and DPF at every iteration as in Fig \ref{fig:stability2}. Fig \ref{fig:stability2} shows the top-1 test accuracy of all iterations throughout one epoch (e.g. 120, 200) during training. In the 120th epoch before the first learning rate decaying, all the methods have a cyclic pattern every 16 iterations due to the pruning frequency of 16. However, there is a clear difference in the scale of performance degradation between DCIL-based methods and DPF. In DPF, the test accuracy decreases drastically by 10-20\% after the pruning mask changes and restores performance before the next pruning mask iteration, whereas DCIL has a stable pattern with much less performance reduction of around 2-5\%. Performance reduction of DCIL without KL is around 3-7\% which is greater than that of DCIL with KL, but the gap is also much smaller than DPF. In the 200th epoch, the difference becomes more pronounced as training progresses using the refined gradient. Although DPF still suffers significant performance degradation each time the pruning mask changes, our DCIL exhibits a much more stable pattern similar to the dense models, because refined gradients make inactive weights as prepared candidates for active weights. The results of structured pruning shows a similar tendency with that of unstructured pruning, which is shown in the Appendix.

\subsection{Cost of training}
One limitation of DCIL in the current implementation is that the model requires two forward and backward paths through the P-Net and the S-Net, taking around 1.5x wall-clock time for training than DPF. 
Nonetheless, with an efficient implementation, the cost of backward paths can be significantly reduced, because essentially the same number of weights (inactive + active) requires update like DPF. Some relevant methods \citep{wei2017minimal, sun2017meprop} utilizing sparse gradients have been proposed to reduce computation in the backward phase. 
Note that the backward path accounts for the majority of the computation requiring around 2-3x more time than the forward path for residual networks \citep{goli2020resprop}.
Most importantly, a better performing model with the same number of parameters can be obtained \textit{for inference} with our method.

Interestingly, even in the current setting, our method achieves on par or superior performance than the SOTA method with equal computation cost due to the faster convergence and training stability.
For instance, in CIFAR-10 ResNet-20 for target sparsity 95\%, DPF achieves 90.34\% with additional finetuning epochs to improve performance (300+60 epochs), whereas ours achieves 90.93\% at 240 epoch, two thirds of the 360 epochs of DPF.

\section{Conclusion}
To get the best out of model efficiency, \emph{dynamic pruning} methods have been studied, which find an efficient sparse network with dynamic sparse patterns. To make diverse sparse patterns, reviving inactive weights with coarse gradients using STE has been used. This causes instability during training and performance degradation due to the gradient approximation. In this work, we propose a novel Dynamic Collective Intelligence Learning (DCIL) which finds a sparse model by training inactive weights with refined gradients rather than using approximated coarse gradients. This can help make inactive weights be superior candidates for future active weights. DCIL outperforms other pruning methods with enhanced stability for various architectures on CIFAR and ImageNet.

\bibliography{aaai22}


\clearpage

\onecolumn

\section{\fontsize{16}{12}\selectfont Supplementary Materials on \\ Dynamic Collective Intelligence Learning: Finding Efficient Sparse Model \\ via Refined Gradients for Pruned Weights}

\vspace{2cm}





\section{Algorithm}

\begin{algorithm*}[h]
\centering

\renewcommand{\algorithmicrequire}{\textbf{Input:}}
\renewcommand{\algorithmicensure}{\textbf{Output:}}
\caption{\text{ Dynamic Collective Intelligence Learning (DCIL)}}
\label{algo-DCIL}
\begin{algorithmic}[1]
\Require Mask update frequency $\mathbf{F}$, Binary mask $\mathbf{M}\in\{0, 1\}^P$, S-Net $\mathbf{W} \in \mathbb{R}^P$, P-Net $\mathbf{\overline{{W}}={M} \odot W}$ \algorithmiccomment{$P$ is the number of parameters and $\odot$ denotes the Hadamard product}
\State \textbf{[ Train ]}
\For{ Iter = 1 ,..., $T$}
\If{Iter $\%$ $\mathbf{F}==0$ } 
\State compute binary mask $\mathbf{M}$ with Eq.(1) and the pruning criterion \algorithmiccomment{Update mask every $\mathbf{F}$ iterations}
\EndIf
\State $\mathbf{W}  \leftarrow \mathbf{W} - \eta\{\mathbf{M} \odot \nabla_{\overline{\mathbf{W}}}\mathcal{L}
            +(1-\mathbf{M}) \odot \nabla_{{\mathbf{W}}}\mathcal{L}\}$ \algorithmiccomment{Update weights which are forwarded with S-Net and P-Net}
\EndFor
\Ensure S-Net, P-Net \algorithmiccomment{P-Net is the efficient network trained from DCIL}
\State \textbf{[Test]}
\State Inference with P-Net

\end{algorithmic}
\end{algorithm*}

In the main paper, we set the mask frequency $\mathbf{F}$ as 16 and use the L2 magnitude for the pruning criterion. Algorithm \ref{algo-DCIL} shows the pseudocode of the proposed DCIL. At every $\mathbf{F}$ iterations, binary mask $\mathbf{M}$ is updated based on Eq.(1) and the pruning criterion. During the update of the dynamic mask, inactive weights which are trained with refined gradients can be altered into active weights. In this process, prepared candidates trained with refined gradients maintain the training stability.

\begin{figure*}[h!]
\centering
  \begin{subfigure}[b]{0.49\textwidth}
    \includegraphics[width=1\textwidth]{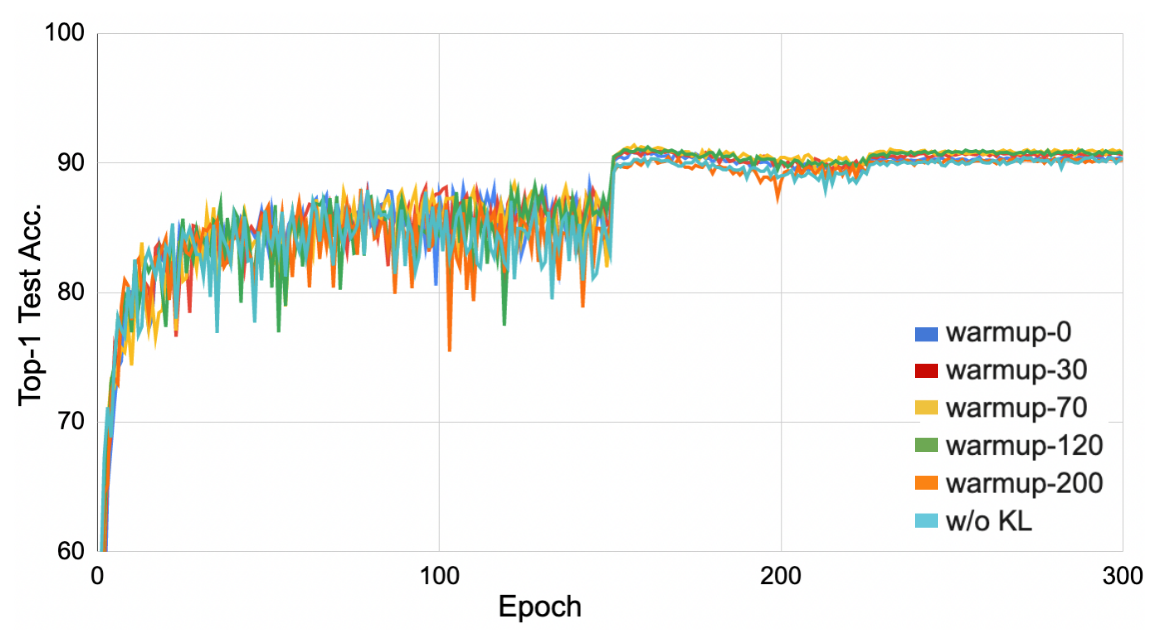}
  \caption{ResNet-20 with 95\% pruning on CIFAR-10}
  \end{subfigure}
  \centering
  \begin{subfigure}[b]{0.49\textwidth}
    \includegraphics[width=1\textwidth]{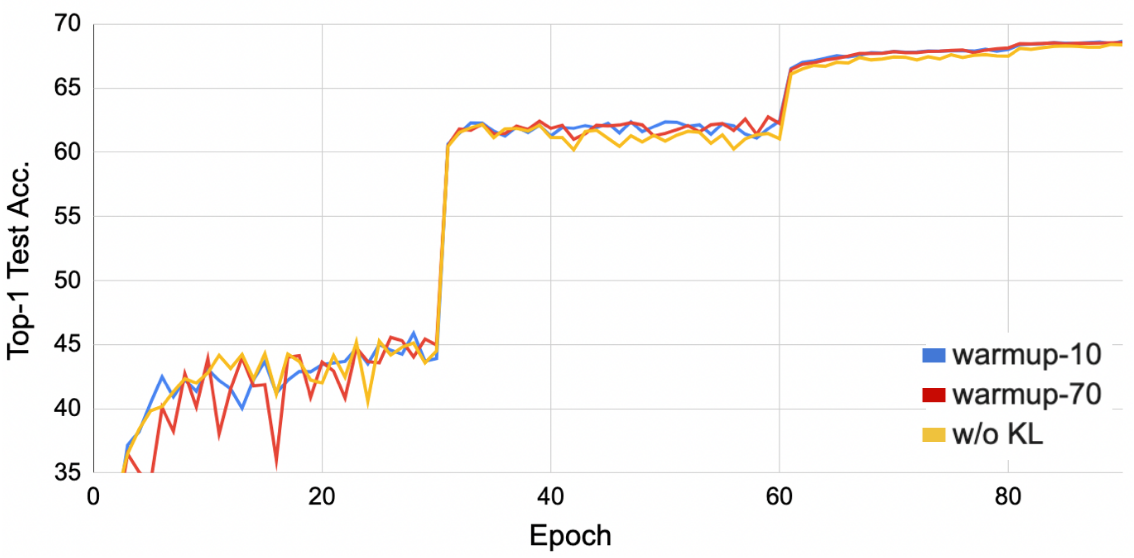}
    \caption{ResNet-18 with 90\% pruning on ImageNet}
  \end{subfigure}
  \caption{Top-1 test accuracy on various warm-up epochs.  }
  \label{appendix:fig:warmup_epoch}
\end{figure*}

\section{Tendency of warm-up }

We introduced the warm-up epoch, an early period of learning without KL loss but only using the cross-entropy loss, and confirmed that the performance is stable across various warm-up epochs. The length of the warm-up epoch does not significantly affect performance and we did not search the hyper-parameter of warm-up in this work. Also, the KL loss help enhance the performance somewhat. 

\subsection{CIFAR10}

Table \ref{appendix:tab:Tendency of Warmup Epoch} shows the top-1 test accuracy for unstructured pruning of 95\% sparsity according to the warm-up epoch in ResNet-20 and ResNet-32 and Fig. \ref{appendix:fig:warmup_epoch}(a) compares the top-1 test accuracy across different warm-up epochs. The performance gap due to the different warm-up epoch is not significant, and the value of 70 which is used in the main paper for warm-up epoch may be tuned.

\begin{table*}[h!]
	\centering
	\caption{ Top-1 test accuracy on CIFAR10 according to warm-up epoch. We report the \emph{last accuracy} for each warm-up epoch value. All reported numbers are averaged over three times} 
	\label{appendix:tab:Tendency of Warmup Epoch}
	\resizebox{1.\textwidth}{!}{%
	{
	\begin{tabular}{lccccccc}
		\toprule
		\parbox{2cm}{\centering Model} & \parbox{2cm}{\centering Target Pr.ratio} & \parbox{2cm}{\centering $w=0$} & \parbox{2cm}{\centering $w=30$} & \parbox{2cm}{\centering $w=70$} & \parbox{2cm}{\centering $w=120$} & \parbox{2cm}{\centering $w=200$} & \parbox{2cm}{\centering $w=300$\\(w/o KL)} \\
		\midrule
		ResNet-20 & 95\% & $90.43 \pm 0.31$ & $90.46 \pm 0.14$ & $90.54 \pm 0.19$ & $90.81 \pm 0.22$ & $90.14 \pm 0.22$ & $90.07 \pm 0.31$ \\
		\midrule
		ResNet-32 & 95\% & $91.85 \pm 0.38$ & $92.02 \pm 0.06$ & $92.04 \pm 0.24$ & $92.01 \pm 0.25$ & $91.92 \pm 0.10$ & $91.41 \pm 0.22$ \\
		\bottomrule
	\end{tabular}%
	}}
\end{table*}


\subsection{ImageNet}
We also applied different warm-up epochs (10, 70 and DCIL w/o KL) on the ImageNet dataset to confirm the effect of warm-up epoch. Table \ref{appendix:tab:Tendency_imagenet} and Fig. \ref{appendix:fig:warmup_epoch}(b) shows the numbers and curves of ResNet-18 on the ImageNet dataset.

\begin{table*}[h!]
	\centering
	\caption{Top-1 test accuracy on ImageNet according to warm-up epoch. We report the \emph{last accuracy} for each warm-up epoch value.} 
	\label{appendix:tab:Tendency_imagenet}
	\resizebox{0.7\textwidth}{!}{%
	{
	\begin{tabular}{lccccccc}
		\toprule
		\parbox{2cm}{\centering Model} & \parbox{2cm}{\centering Target Pr.ratio} &  \parbox{2cm}{\centering $w=10$} & \parbox{2cm}{\centering $w=70$} &  \parbox{2cm}{\centering $w=90$\\(w/o KL)} \\
		\midrule
		ResNet-18 & 90\% & $68.66$ & $68.55$ & $68.37$  \\
		\bottomrule
	\end{tabular}%
	}}
\end{table*}


\section{More detailed experiments}

\subsection{Differences between Dense and pruned model}
In this section, we provide the performance differences between the dense model and the pruned model (`Pruned - Dense') to compare the degree of the performance degradation. Table \ref{appendix:diff_unstrcuted} and Table \ref{appendix:diff_strcuted}  show the performance degradation according to the experiments in the main paper.

\begin{table*}[h!]
	\centering
	\caption{Top-1 test accuracy \emph{differences} (`Pruned - Dense') of our DCIL and other baseline methods on CIFAR-10 for unstructured pruning. $\star$ means that the model does not converge. $^\dagger$ indicates our numbers while the unmarked others are calculated directly from the DPF paper. We ran DPF$^\dagger$ using the official code and official training schedule and all reported numbers are averaged over three times.}
	\label{appendix:diff_unstrcuted}
	\resizebox{1.\textwidth}{!}{%
	{
	\begin{tabular}{lcccccccccc}
		\toprule
		\multirow{2}{*}{\parbox{2cm}{\centering Model}} & \multirow{2}{*}{\parbox{2cm}{\centering Dense}} & \multirow{2}{*}{\parbox{1.5cm}{\centering SM\\ (DZ,\citeyear{dettmers2019sparse})}} & \multirow{2}{*}{\parbox{1.5cm}{\centering DSR\\ (MW,\citeyear{mostafa2019parameter})}} & \multirow{2}{*}{\parbox{1.5cm}{\centering DPF\\ (Lin,\citeyear{lin2020dynamic}}} & \multirow{2}{*}{\parbox{2cm}{\centering Dense$^\dagger$}} & \multicolumn{2}{c}{\centering DPF$^\dagger$} & \multicolumn{2}{c}{\centering DCIL (Ours)} & \multirow{2}{*}{\parbox{2cm}{\centering Target Pr.\\ ratio}} \\
		\cmidrule(lr){7-8} \cmidrule(lr){9-10}
		& & & & & & \centering Last & \centering Best & \centering Last & \centering Best & \\
		\midrule
		\multirow{2}{*}{ResNet-20} & \multirow{2}{*}{$92.48 \pm 0.20$} & $-2.72$ & $-4.60$ & $-1.6$ & \multirow{2}{*}{$92.36 \pm 0.10$} & $-4.34$ & $-1.49$ & $\mathbf{-0.75}$ & $-0.43$ & 90\% \\
		& & $-9.45$ & $\star$ & $-4.47$ & & $-10.9$ & $-4.52$ & $\mathbf{-1.82}$ & $-1.56$ & 95\% \\
		\midrule
		\multirow{2}{*}{ResNet-32} & \multirow{2}{*}{$93.83 \pm 0.12$} & $-2.29$ & $-2.42$ & $-1.41$ & \multirow{2}{*}{$93.22 \pm 0.07$} & $-2.08$ & $-0.83$ & $\mathbf{-0.17}$ & $0.01$ & 90\% \\
		& & $-5.15$ & $-9.71$ & $-2.89$ & & $-6.7$ & $-2.31$ & $\mathbf{-1.18}$ & $-0.95$ &  95\% \\
		\midrule
		\multirow{2}{*}{ResNet-56} & \multirow{2}{*}{$94.51 \pm 0.20$} & $-1.78$ & $-0.73$ & $-0.56$ & \multirow{2}{*}{$94.34 \pm 0.19$} & $-0.72$ & $-0.37$ & $\mathbf{-0.18}$ & $0.05$ & 90\% \\
		& & $-3.55$ & $-1.94$ & $-1.77$ & & $-3.75$ & $-1.53$ & $\mathbf{-0.59}$ & $-0.36$ & 95\% \\
		\midrule
		\multirow{3}{*}{WideResNet-28-2} & \multirow{3}{*}{$95.01 \pm 0.04$} & $-1.6$ & $-1.13$ & $-0.65$ & \multirow{3}{*}{$94.73 \pm 0.03$} & $-0.65$ & $-0.41$ & $\mathbf{-0.04}$ & $0.11$ & 90\% \\
		& & $-2.77$ & $-2.27$ & $-1.39$ & & $-1.6$ & $-1.12$ & $\mathbf{-0.72}$ & $-0.54$ & 95\% \\
		& & $-9.65$ & $\star$ & $-6.09$ & & $-8.91$ & $-5.96$ & $\mathbf{-3.54}$ & $-3.38$ & 99\% \\
		\bottomrule
	\end{tabular}%
	}}
\end{table*}

\begin{table*}[h!]
	\centering
	\caption{Top-1 test accuracy \emph{differences} (`Pruned - Dense') of our DCIL and other baseline methods on CIFAR-10 for structured pruning. The settings and symbols are the same as in Table \ref{appendix:diff_unstrcuted}}
	\label{appendix:diff_strcuted}
	\resizebox{1.\textwidth}{!}{%
	{
	\begin{tabular}{lccccccccc}
		\toprule
		\multirow{2}{*}{\parbox{2cm}{\centering Model}} & \multirow{2}{*}{\parbox{2cm}{\centering Dense}} & \multirow{2}{*}{\parbox{1.5cm}{\centering SFP\\ (He,\citeyear{he2018soft})}} & \multirow{2}{*}{\parbox{1.5cm}{\centering DPF\\ (Lin,\citeyear{lin2020dynamic})}} & \multirow{2}{*}{\parbox{2cm}{\centering Dense$^\dagger$}} & \multicolumn{2}{c}{\centering DPF$^\dagger$} & \multicolumn{2}{c}{\centering DCIL (Ours)} & \multirow{2}{*}{\parbox{2cm}{\centering Target Pr.\\ ratio}} \\
		\cmidrule(lr){6-7} \cmidrule(lr){8-9}
		& & & & & \centering Last & \centering Best & \centering Last & \centering Best & \\
		\midrule
		\multirow{2}{*}{ResNet-32} & \multirow{2}{*}{$93.52 \pm 0.13$} & $-1.34$ & $-1.45$ & \multirow{2}{*}{$93.22 \pm 0.07$} & $-1.61$ & $-1.40$ & $\mathbf{-0.19}$ & $-0.12$ & 30\% \\
		& & $-2.38$ & $-2.02$ & & $-3.20$ & $-2.93$ & $\mathbf{-0.61}$ & $-0.4$ & 40\% \\
		\midrule
		\multirow{2}{*}{WideResNet-28-2} & \multirow{2}{*}{$95.01 \pm 0.04$} & $-0.99$ & $-0.49$ & \multirow{2}{*}{$94.73 \pm 0.03$} & $-0.63$ & $-0.43$ & $\mathbf{-0.19}$ & $-0.13$ & 40\% \\
		& & $-9.01$ & $-4.48$ & & $-0.91$ & $-0.74$ & $\mathbf{-0.52}$ & $-0.43$ &  80\% \\
		\midrule
		\multirow{2}{*}{WideResNet-28-4} & \multirow{2}{*}{$95.69 \pm 0.10$} & $-0.54$ & $-0.19$ & \multirow{2}{*}{$95.50 \pm 0.07$} & $-0.14$ & $0.01$ & ${-0.04}$ & $0.09$ & 40\% \\
		& & $-3.81$ & $-1.9$ & & $-2.09$ & $-1.99$ & $\mathbf{-1.32}$ & $-1.13$ & 80\% \\
		\midrule
		\multirow{2}{*}{WideResNet-28-8} & \multirow{2}{*}{$96.06 \pm 0.06$} & $-0.44$ & $0.00$ & \multirow{2}{*}{$95.93 \pm 0.07$} & $-0.06$ & ${0.10}$ & $-0.20$ & $-0.01$ & 40\% \\
		& & $-1.84$ & $-0.91$ & & $-0.96$ & $-0.77$ & $\mathbf{-0.50}$ & $-0.44$ & 80\% \\
		\bottomrule
	\end{tabular}%
	}}
\end{table*}

\subsection{Structured pruning with CIFAR-100}

We compared our DCIL with other structured pruning methods on CIFAR-100. The details of structured pruning is same as our main paper. Table \ref{appendix:tab:CIFAR100_structured} shows the numbers of structured pruning experiments on CIFAR-100.

\begin{table*}[h!]
	\centering
	\caption{Top-1 test accuracy of our DCIL and other baseline methods on CIFAR-100 for structured weight pruning. The same training scheme was used as on CIFAR-10. To compare the stability of pruning methods we reported the accuracy of the last epoch and the best accuracy of DCIL.}
	\label{appendix:tab:CIFAR100_structured}
	\resizebox{1.\textwidth}{!}{%
	{\Large
	\begin{tabular}{lccccccccccc}
		\toprule
		\multirow{2}{*}{\parbox{2cm}{\centering Model}} & \multicolumn{4}{c}{\centering DCUL (Ours)} & \multicolumn{2}{c}{\parbox{2.5cm}{\centering SFP \\ (He, \citeyear{he2018soft})}} & \multicolumn{2}{c}{\parbox{2.5cm}{\centering MIL \\ (Dong, \citeyear{dong2017less})}} & \multicolumn{2}{c}{\parbox{2.5cm}{\centering FPGM \\ (He, \citeyear{he2019filter})}} &
		\multirow{2}{*}{\parbox{2cm}{\centering Target Pr.\\ ratio}} \\
		\cmidrule(lr){2-5} \cmidrule(lr){6-7} \cmidrule(lr){8-9} \cmidrule(lr){10-11}
		& \centering Dense & \centering Last & \centering Best & \centering Diff. & \centering Dense & \centering Diff. & \centering Dense & \centering Diff. & \centering Dense & \centering Diff. & \\
		\midrule
		\multirow{2}{*}{ResNet-56} & \multirow{2}{*}{$74.23 \pm 0.13$} & $73.49 \pm 0.16$ & $ 73.80 \pm 0.23$ & $\mathbf{-0.74}$ & \multirow{2}{*}{$71.4$} & $-$ & \multirow{2}{*}{$71.33$} & $-2.96$ & \multirow{2}{*}{$71.41$} & $-$ & 40\% \\
		& & $73.17 \pm 0.04$ & $73.59 \pm 0.15$ & $\mathbf{-1.06}$ & & $-2.61$ & & $-$ & & $-1.75$ & 50\% \\
		\midrule
	\end{tabular}%
	}}
\end{table*}

\section{Additional Analysis of the stability}

\begin{figure*}[h!]
\centering
  \includegraphics[width=0.6\textwidth]{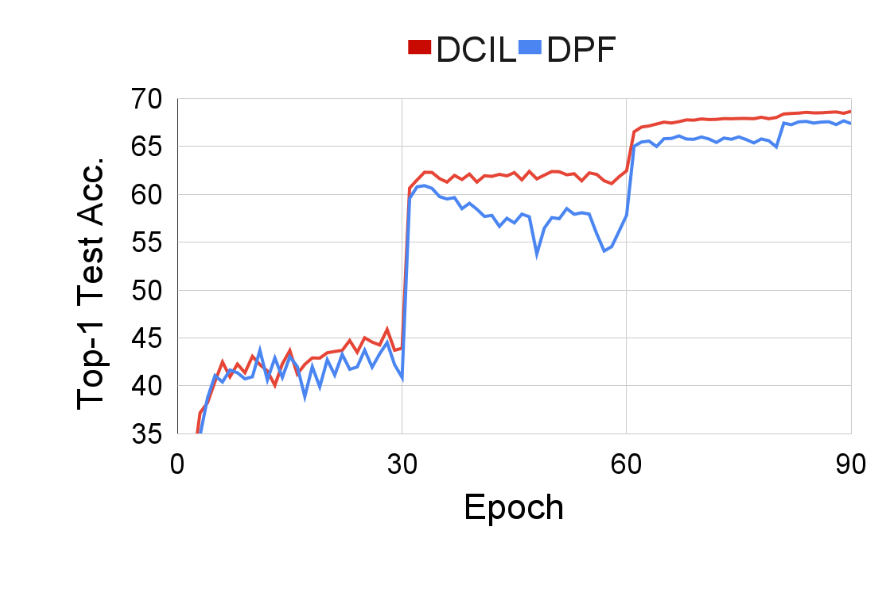}
  \caption{Training stability, unstructured pruning with 90\%, ResNet-18 on Imagenet.}
  \label{appendix:fig:stability_unstructured_imagenet}
\end{figure*}

\begin{figure*}[h!]
\centering
  \begin{subfigure}[b]{0.49\textwidth}
    \includegraphics[width=1\textwidth]{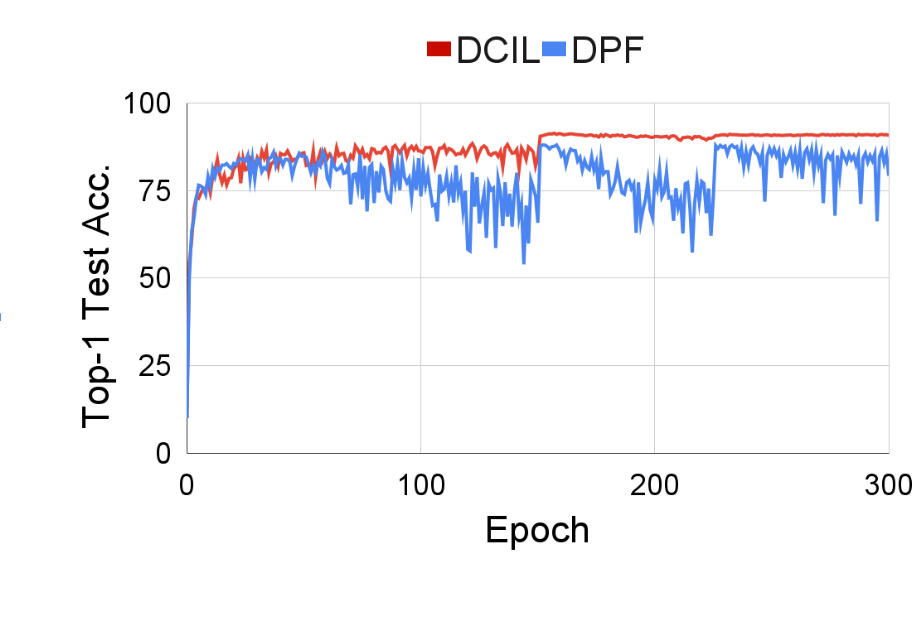}
  \caption{Unstructured pruning}
  \end{subfigure}
  \centering
  \begin{subfigure}[b]{0.49\textwidth}
    \includegraphics[width=1\textwidth]{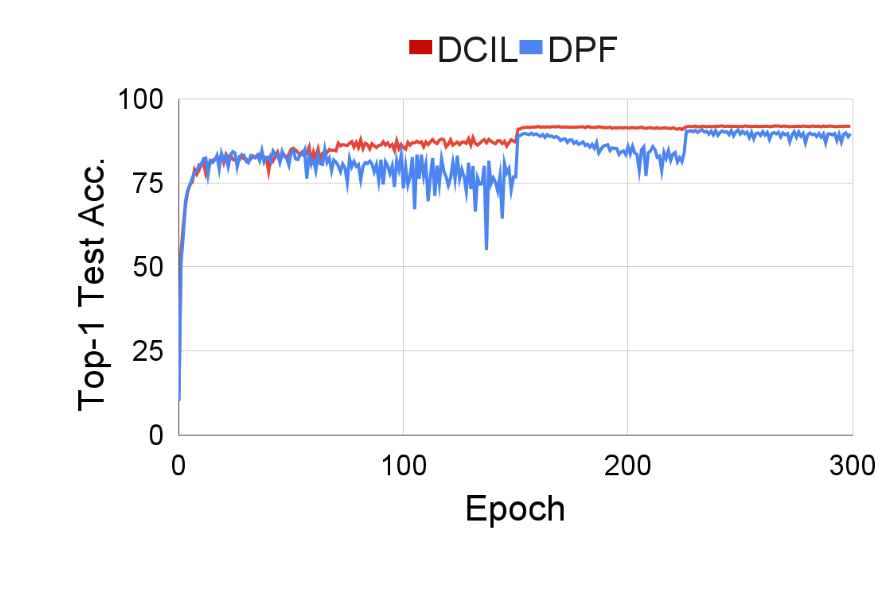}
    \caption{Structured pruning}
  \end{subfigure}
  \caption{Top-1 test accuracy vs. epoch to show the training stability. ResNet-20 with 95\% pruning on CIFAR-10 dataset, using unstructured and structured pruning.}
  \label{appendix:fig:stability2}
\end{figure*}

As mentioned in the main paper, our DCIL has an advantage in the training stability. We provide the overall Top-1 test accuracy vs epoch in Fig. \ref{appendix:fig:stability_unstructured_imagenet} and Fig. \ref{appendix:fig:stability2}.


\section{Fast convergence of DCIL}
In the \textit{Cost of training} subsection of the main paper, we noted that the limitation of DCIL is that the model takes around 1.5x wall-clock time for training than DPF. To address this limitation, we suggested that training could be stopped earlier due to the training stability, citing that the accuracy of DCIL at the 240th epoch is higher than that of DPF which is fine-tuned for 60 epochs additionally. Moreover, we can finish the training much earlier by adjusting the target epoch, the epoch at which the sparsity of the model reaches the target sparsity. We additionally conducted experiments by adjusting the target epochs smaller than 225 used in the main paper, and the learning rate decaying epochs were also adjusted so that the second learning rate decaying epochs are the same as the target epochs.
Table \ref{appendix:tab:Fast Convergence} compares the accuracy and training time between DCIL with reduced training time and DPF which is fine-tuned. For DCIL, we reduced the target epoch to 100, 125, 150, 175, 200 and trained only 25 epochs additional after that target epoch. Since DCIL is much more stable in training, it has higher performance with only 0.6-0.7 times the training time compared to DPF.

\begin{table*}[h!]
	\centering
	\caption{Top-1 test accuracy on CIFAR10 dataset in unstructured pruning of target sparsity 95\%. For DCIL, we reduced the training time by adjusting the target epoch and training only 25 additional epochs after the target epoch. For DPF, fine-tuning for 60 epochs is required to improve performance. We compare the accuracy and training time of DCIL with reduced training time to both original DPF and fine-tuned DPF in ResNet-20 and ResNet-32. \textit{Time} column refers to the ratio compared to the DPF training and fine-tuning time. All numbers of DPF are from the original paper of DPF \cite{lin2020dynamic}.} 
	\label{appendix:tab:Fast Convergence}
	\resizebox{0.7\textwidth}{!}{%
	{
	\begin{tabular}{lccccc}
		\toprule
		\multirow{2}{*}{\parbox{2cm}{\centering Method}} & \multirow{2}{*}{\parbox{2cm}{\centering Total Epoch}} & \multirow{2}{*}{\parbox{2cm}{\centering Target Epoch}} & \multicolumn{2}{c}{\parbox{4cm}{\centering Top-1 Test Acc.}} & \multirow{2}{*}{\parbox{2cm}{\centering Time}} \\		
		\cmidrule(lr){4-5}
		& & & \centering ResNet-20 & \centering ResNet-32 & \\
		\midrule
		\multirow{6}{*}{\parbox{2cm}{\centering DCIL}} & $125$ & $100$ & $89.63$ & $92.07$ & $\times 0.52$  \\
		& $150$ & $125$ & $90.04$ & $92.56$ & $\times 0.63$  \\
		& $175$ & $150$ & $90.50$ & $92.93$ & $\times 0.73$  \\
		& $200$ & $175$ & $90.62$ & $93.15$ & $\times 0.83$  \\
		& $225$ & $200$ & $90.49$ & $93.03$ & $\times 0.94$  \\
		& $250$ & $225$ & $91.06$ & $93.23$ & $\times 1.04$  \\
		\midrule
		\multirow{2}{*}{\parbox{2cm}{\centering DPF}} & $300$ & $225$ & $88.01$ & $90.94$ & $\times 0.83$ \\
		& $360\ (300 + 60)$ & $225$ & $90.34$ & $92.18$ & $\times 1.0$ \\
		\bottomrule
	\end{tabular}%
	}}
\end{table*}


\end{document}